\documentclass[10pt,conference]{IEEEtran}
\usepackage{cite}
\usepackage{amsmath,amssymb,amsfonts}
\usepackage{algorithmic}
\usepackage{graphicx}
\usepackage{textcomp}
\usepackage{xcolor}
\def\BibTeX{{\rm B\kern-.05em{\sc i\kern-.025em b}\kern-.08em
  T\kern-.1667em\lower.7ex\hbox{E}\kern-.125emX}}

\usepackage[pdftex,
            pdfauthor={Elizabeth Ditton, Anne Swinbourne, Trina Myers, and Mitchell Scovell},
            pdftitle={Applying Semi-Automated Hyperparameter Tuning for Clustering Algorithms},
            pdfsubject={Applying Semi-Automated Hyperparameter Tuning for Clustering Algorithms},
            pdfkeywords={Machine Learning, Clustering Algorithms, Hyperparameter Tuning, Persona Development},
            pdfproducer={},
            pdfcreator={}]{hyperref}

\begin{document}

\title{Applying Semi-Automated Hyperparameter Tuning for Clustering Algorithms}

\author{
\IEEEauthorblockN{
Elizabeth Ditton\IEEEauthorrefmark{1}\IEEEauthorrefmark{2},
Anne Swinbourne\IEEEauthorrefmark{2},
Trina Myers\IEEEauthorrefmark{3},
and Mitchell Scovell\IEEEauthorrefmark{2}}

\IEEEauthorblockA{\IEEEauthorrefmark{2}James Cook University, Townsville, Australia}
\IEEEauthorblockA{\IEEEauthorrefmark{3}Queensland University of Technology, Brisbane, Australia}
\IEEEauthorblockA{\IEEEauthorrefmark{1}Email: elizabeth.forest@my.jcu.edu.au}
}

\maketitle

\begin{abstract}

When approaching a clustering problem, choosing the right clustering algorithm and parameters is essential, as each clustering algorithm is proficient at finding clusters of a particular nature.
Due to the unsupervised nature of clustering algorithms, there are no ground truth values available for empirical evaluation, which makes automation of the parameter selection process through hyperparameter tuning difficult.
Previous approaches to hyperparameter tuning for clustering algorithms have relied on internal metrics, which are often biased towards certain algorithms, or having some ground truth labels available, moving the problem into the semi-supervised space.
This preliminary study proposes a framework for semi-automated hyperparameter tuning of clustering problems, using a grid search to develop a series of graphs and easy to interpret metrics that can then be used for more efficient domain-specific evaluation.
Preliminary results show that internal metrics are unable to capture the semantic quality of the clusters developed and approaches driven by internal metrics would come to different conclusions than those driven by manual evaluation.

\end{abstract}


\section{Introduction}

Clustering is an area of unsupervised machine learning that attempts to find structure in unstructured data by creating groups of similar values \cite{jain_data_2010, xu_comprehensive_2015}.
One of the primary challenges of clustering is that there are numerous algorithms, and algorithm selection can have a drastic impact on performance. Furthermore, the performance of a particular algorithm is often dependent on the nature of the clusters in the data \cite{jain_data_2010}.
Even two similar algorithms may find completely different sets of clusters in the same data set \cite{jain_data_2010}.
Clustering algorithms are also notoriously difficult to evaluate, as there is no ground truth available and multiple sets of clusters created from one data set could be equally valid \cite{jain_data_2010}.

The selection of a clustering algorithm and the algorithm parameters, a process known as hyperparameter tuning, is a considerable challenge when applying a clustering solution to real-world problems. Multiple iterations and considerable domain knowledge is often required to find an optimal algorithm configuration, and the process is often long and tedious \cite{fan_hyperparameter_2020, van_craenendonck_constraint-based_2017}. In supervised problems, where a ground truth is available, hyperparameter tuning is often automated, however, automated hyperparameter tuning requires accurate and objective evaluation metrics. As evaluating clustering algorithms is a considerable problem, completely automated methods of hyperparameter tuning for clustering algorithms often rely on internal evaluation metrics \cite{fan_hyperparameter_2020, blumenberg_hypercluster_2020, shalamov_reinforcement-based_2018}, or having some ground truth labels available for external evaluation metrics \cite{van_craenendonck_constraint-based_2017, minku_novel_2019}, which moves the problem into the semi-automated space.

However, these methods of evaluation are often flawed, and cannot comment on the quality of the clusters developed for the use case \cite{von_luxburg_clustering_2012}. Internal methods measure the cluster quality with similarity metrics and tend to be biased towards particular types of clustering algorithms \cite{von_luxburg_clustering_2012}. Another method of evaluation is through meta-criteria, such as stability and statistical significance, which can be useful in determining the quality of a clustering algorithm but less so in comparing the results of multiple algorithms. Von Luxburg et al. \cite{von_luxburg_clustering_2012} asserted that clustering algorithms cannot be evaluated independently to the context in which they will be used.
Domain specific evaluation can be highly subjective and often requires significant time and resources to perform.
As the effect of hyperparameters on clustering results cannot be described through a convex function, an exhaustive grid search is required to find the optimal hyperparameters \cite{blumenberg_hypercluster_2020}. For an individual to manually perform an exhaustive grid search and evaluate all of the possible results would be a time intensive and cumbersome process.

We propose a framework for semi-automated hyperparameter tuning of clustering problems, using internal metrics and meta-criteria to guide an individual performing manual, domain specific evaluation. Preliminary results were found by running the framework to identify the most appropriate algorithm and parameter combination for persona development. The results illustrated the framework's facilitation of domain specific evaluation and ability to identify more use case relevant results than methods based purely on internal metrics.
The key contributions of this preliminary study are that a framework for the semi-automated hyperparameter tuning of a clustering problem is presented and evaluated on a real-world clustering problem. This is then compared to results using internal metrics for hyperparameter tuning.

\begin{figure*}[!t]
 \centering
 \includegraphics[width=\textwidth]{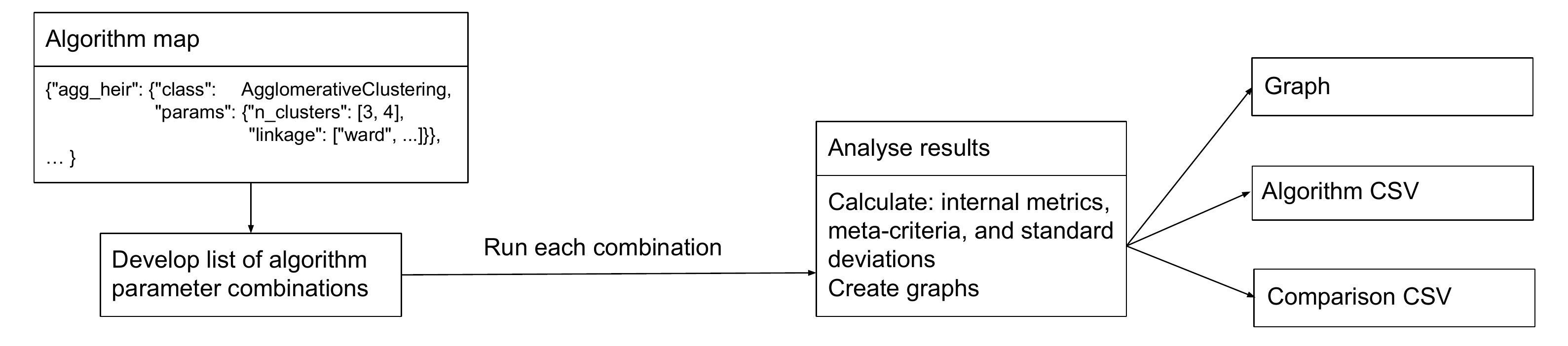}
 \caption{A graphical representation of the automated portion of the semi-automated hyperparameter tuning framework for clustering problems}
 \label{fig:framework}
\end{figure*}

\section{Framework}

The proposed framework performs an exhaustive grid search across multiple clustering algorithms and parameters. The results are then outputted as a set of graphs and simple meta-criteria metrics that can be used for focused domain specific evaluation. An overview of the framework is given in Fig.~\ref{fig:framework}.

\subsection{Grid Search}

The framework takes a map with an identifier as the key and an exhaustive parameter map as the value. The parameter map also gives the function or class used to run the clustering algorithm. Each parameter combination is assigned a unique identifier that is used throughout the output, made up of the identifier given in the map and a number, e.g., kmeans\_v0.

\subsection{Automated Outputs}

A number of metrics are collected from the clusters developed by each parameter combination: cluster sizes; internal metric, specifically Silhouette Coefficient \cite{rousseeuw_silhouettes_1987}, Calinski-Harabasz Index \cite{calinski_dendrite_1974}; and, Davies-Bouldin Index \cite{davies_cluster_1979}; the mean value for each feature in each cluster with the number of standard deviations each cluster mean is from the population mean, and its statistical significance, or p-value. Any features found to be statistically significant are tracked.
All of the data is outputted to a CSV file for the parameter combination, as well as values such as the internal metrics and meta-criteria being additionally outputted to running CSV files for quick reference.
A series of graphs are then built so that each graph represents how many standard deviations a cluster centroid is from the population mean for each of the predefined key features for domain specific evaluation.

\subsection{Domain Specific Evaluation}

When performing the manual evaluation, the individual is encouraged to first use the meta-criteria and internal metrics to rule out unacceptable cluster sets. For example, a set of clusters that has no significant features would be considered unacceptable. The individual can then use the graphs and knowledge of the statistically significant features for the remaining options to perform a subjective, domain specific evaluation. It was found most effective to perform a quick first pass of the graphs to find graphs that showed particularly weak clusters or obviously went against the domain specific evaluation criteria.

\section{Preliminary results}

The framework was used to compare three algorithms for the purpose of persona development based on cyclone preparatory behaviour. A persona is a description of a fictitious person used to describe analytical data and customer segments in a manner that emphasises human attributes and empathy \cite{salminen_are_2018}. Personas are used in a wide range of fields, but primarily for marketing and design purposes. The three algorithms compared, each with multiple parameter options, were:
1) k-means \cite{ball_isodata_1965,lloyd_least_1982,macqueen_methods_1967,steinhaus_sur_1956}; 2) Agglomerative Hierarchical Clustering (AHC); and, 3) Non-negative Matrix Factorization (NMF) \cite{lee_learning_1999}.
These algorithms were selected as they are most common within persona development \cite{salminen_survey_2021}.
The domain specific evaluation performed was based on how well the clusters could be explained via a behavioural model, specifically the Protection Action Decision Model (PADM) \cite{lindell_behavioral_1992,lindell_protective_2012,terpstra_citizens_2013}. The data used was survey data from 519 residents of the cyclone prone North Queensland, Australia \cite{scovell_north_2018-1}.

Of the 16 parameter combinations used, six could be immediately ruled out due to the meta-criteria and a further four were able to be easily ruled out from the graphs, which left six for domain specific evaluation. That is, the framework facilitated the identification of a preferred algorithm and parameter combination, in this case AHC using Ward's linkage and 3 clusters.
This result contradicted what would have been found using a fully automated framework based on internal metrics, as all of the internal metrics preferred other combinations, including combinations which were ruled out by meta-criteria in some cases.

\section{Conclusion and Future Work}

The quality of a set of clusters is highly dependent on the algorithm and parameters used to develop them. However, the subjective nature of cluster evaluation makes hyperparameter tuning difficult to automate, resulting in a time consuming, tedious process.
Previous approaches have relied on having some ground truth labels available, moving the problem out of the unsupervised space, or on internal metrics, which are known to be biased and unreliable.

This preliminary study presented a semi-automated framework for hyperparameter tuning for clustering problems. The framework performs an exhaustive grid search of all algorithm parameter combinations to produce a series of graphs and easy to interpret outputs. Preliminary results show that these graphs and outputs can then be used for efficient domain specific evaluation that can produce results more relevant to the cluster's use case.

\IEEEtriggeratref{11}
\bibliographystyle{IEEEtran}
\bibliography{IEEEabrv,full-lib}

\begin{thebibliography}{10}
\providecommand{\url}[1]{#1}
\csname url@samestyle\endcsname
\providecommand{\newblock}{\relax}
\providecommand{\bibinfo}[2]{#2}
\providecommand{\BIBentrySTDinterwordspacing}{\spaceskip=0pt\relax}
\providecommand{\BIBentryALTinterwordstretchfactor}{4}
\providecommand{\BIBentryALTinterwordspacing}{\spaceskip=\fontdimen2\font plus
\BIBentryALTinterwordstretchfactor\fontdimen3\font minus
  \fontdimen4\font\relax}
\providecommand{\BIBforeignlanguage}[2]{{%
\expandafter\ifx\csname l@#1\endcsname\relax
\typeout{** WARNING: IEEEtran.bst: No hyphenation pattern has been}%
\typeout{** loaded for the language `#1'. Using the pattern for}%
\typeout{** the default language instead.}%
\else
\language=\csname l@#1\endcsname
\fi
#2}}
\providecommand{\BIBdecl}{\relax}
\BIBdecl

\bibitem{jain_data_2010}
\BIBentryALTinterwordspacing
A.~K. Jain, ``Data clustering: 50 years beyond {K}-means,'' \emph{Pattern
  Recognition Letters}, vol.~31, no.~8, pp. 651--666, Jun. 2010. [Online].
  Available:
  \url{http://www.sciencedirect.com/science/article/pii/S0167865509002323}
\BIBentrySTDinterwordspacing

\bibitem{xu_comprehensive_2015}
\BIBentryALTinterwordspacing
D.~Xu and Y.~Tian, ``\BIBforeignlanguage{en}{A comprehensive survey of
  clustering algorithms},'' \emph{\BIBforeignlanguage{en}{Annals of Data
  Science}}, vol.~2, no.~2, pp. 165--193, Jun. 2015. [Online]. Available:
  \url{https://doi.org/10.1007/s40745-015-0040-1}
\BIBentrySTDinterwordspacing

\bibitem{fan_hyperparameter_2020}
\BIBentryALTinterwordspacing
X.~Fan, Y.~Yue, P.~Sarkar, and Y.~X.~R. Wang, ``\BIBforeignlanguage{en}{On
  hyperparameter tuning in general clustering problems},'' in
  \emph{\BIBforeignlanguage{en}{Proceedings of the 37th {International}
  {Conference} on {Machine} {Learning}}}, ser. Proceedings of {Machine}
  {Learning} {Research}, vol. 119.\hskip 1em plus 0.5em minus 0.4em\relax PLMR,
  Jul. 2020, pp. 2996--3007. [Online]. Available:
  \url{http://proceedings.mlr.press/v119/fan20b.html}
\BIBentrySTDinterwordspacing

\bibitem{van_craenendonck_constraint-based_2017}
\BIBentryALTinterwordspacing
T.~Van~Craenendonck and H.~Blockeel, ``\BIBforeignlanguage{en}{Constraint-based
  clustering selection},'' \emph{\BIBforeignlanguage{en}{Machine Learning}},
  vol. 106, no.~9, pp. 1497--1521, Oct. 2017. [Online]. Available:
  \url{https://doi.org/10.1007/s10994-017-5643-7}
\BIBentrySTDinterwordspacing

\bibitem{blumenberg_hypercluster_2020}
\BIBentryALTinterwordspacing
L.~Blumenberg and K.~V. Ruggles, ``\BIBforeignlanguage{en}{Hypercluster: a
  flexible tool for parallelized unsupervised clustering optimization},''
  \emph{\BIBforeignlanguage{en}{BMC Bioinformatics}}, vol.~21, no.~1, p. 428,
  Sep. 2020. [Online]. Available:
  \url{https://doi.org/10.1186/s12859-020-03774-1}
\BIBentrySTDinterwordspacing

\bibitem{shalamov_reinforcement-based_2018}
\BIBentryALTinterwordspacing
V.~Shalamov, V.~Efimova, S.~Muravyov, and A.~Filchenkov,
  ``\BIBforeignlanguage{en}{Reinforcement-based {Method} for {Simultaneous}
  {Clustering} {Algorithm} {Selection} and its {Hyperparameters}
  {Optimization}},'' \emph{\BIBforeignlanguage{en}{Procedia Computer Science}},
  vol. 136, pp. 144--153, Jan. 2018, publisher: Elsevier. [Online]. Available:
  \url{http://www.sciencedirect.com/science/article/pii/S1877050918315527}
\BIBentrySTDinterwordspacing

\bibitem{minku_novel_2019}
\BIBentryALTinterwordspacing
L.~L. Minku, ``\BIBforeignlanguage{en}{A novel online supervised hyperparameter
  tuning procedure applied to cross-company software effort estimation},''
  \emph{\BIBforeignlanguage{en}{Empirical Software Engineering}}, vol.~24,
  no.~5, pp. 3153--3204, Oct. 2019. [Online]. Available:
  \url{https://doi.org/10.1007/s10664-019-09686-w}
\BIBentrySTDinterwordspacing

\bibitem{von_luxburg_clustering_2012}
U.~Von~Luxburg, R.~C. Williamson, and I.~Guyon, ``Clustering: {Science} or
  art?'' 2012, pp. 65--79.

\bibitem{rousseeuw_silhouettes_1987}
\BIBentryALTinterwordspacing
P.~J. Rousseeuw, ``\BIBforeignlanguage{en}{Silhouettes: {A} graphical aid to
  the interpretation and validation of cluster analysis},''
  \emph{\BIBforeignlanguage{en}{Journal of Computational and Applied
  Mathematics}}, vol.~20, pp. 53--65, Nov. 1987, publisher: North-Holland.
  [Online]. Available:
  \url{http://www.sciencedirect.com/science/article/pii/0377042787901257}
\BIBentrySTDinterwordspacing

\bibitem{calinski_dendrite_1974}
T.~Caliński and J.~Harabasz, ``A {Dendrite} {Method} for {Cluster}
  {Analysis},'' \emph{Communications in Statistics - Theory and Methods},
  vol.~3, pp. 1--27, Jan. 1974.

\bibitem{davies_cluster_1979}
D.~L. Davies and D.~W. Bouldin, ``A {Cluster} {Separation} {Measure},''
  \emph{IEEE Transactions on Pattern Analysis and Machine Intelligence}, vol.
  PAMI-1, no.~2, pp. 224--227, Apr. 1979, conference Name: IEEE Transactions on
  Pattern Analysis and Machine Intelligence.

\bibitem{salminen_are_2018}
\BIBentryALTinterwordspacing
J.~Salminen, B.~J. Jansen, J.~An, H.~Kwak, and S.-g. Jung,
  ``\BIBforeignlanguage{en}{Are personas done? {Evaluating} their usefulness in
  the age of digital analytics},'' \emph{\BIBforeignlanguage{en}{Persona
  Studies}}, vol.~4, no.~2, pp. 47--65, Nov. 2018. [Online]. Available:
  \url{https://ojs.deakin.edu.au/index.php/ps/article/view/737}
\BIBentrySTDinterwordspacing

\bibitem{ball_isodata_1965}
G.~H. Ball and D.~J. Hall, ``{ISODATA}, a novel method of data analysis and
  pattern classification,'' Stanford research inst Menlo Park CA, Tech. Rep.,
  1965.

\bibitem{lloyd_least_1982}
S.~Lloyd, ``Least squares quantization in {PCM},'' \emph{IEEE transactions on
  information theory}, vol.~28, no.~2, pp. 129--137, 1982.

\bibitem{macqueen_methods_1967}
J.~MacQueen, ``Some methods for classification and analysis of multivariate
  observations,'' in \emph{Proceedings of the fifth {Berkeley} symposium on
  mathematical statistics and probability}, vol.~1.\hskip 1em plus 0.5em minus
  0.4em\relax Oakland, CA, USA, 1967, pp. 281--297.

\bibitem{steinhaus_sur_1956}
H.~Steinhaus, ``Sur la division des corp materiels en parties,'' \emph{Bull.
  Acad. Polon. Sci}, vol.~1, no. 804, p. 801, 1956.

\bibitem{lee_learning_1999}
D.~D. Lee and H.~S. Seung, ``Learning the parts of objects by non-negative
  matrix factorization,'' \emph{Nature}, vol. 401, no. 6755, pp. 788--791,
  1999, publisher: Nature Publishing Group.

\bibitem{salminen_survey_2021}
\BIBentryALTinterwordspacing
J.~Salminen, K.~Guan, S.-G. Jung, and B.~J. Jansen, ``A survey of 15 years of
  data-driven persona development,'' \emph{International Journal of
  Human–Computer Interaction}, vol.~0, no.~0, pp. 1--24, 2021. [Online].
  Available: \url{https://doi.org/10.1080/10447318.2021.1908670}
\BIBentrySTDinterwordspacing

\bibitem{lindell_behavioral_1992}
M.~K. Lindell and R.~W. Perry, \emph{Behavioral foundations of community
  emergency planning}, ser. Behavioral foundations of community emergency
  planning.\hskip 1em plus 0.5em minus 0.4em\relax Washington, DC, US:
  Hemisphere Publishing Corp, 1992.

\bibitem{lindell_protective_2012}
------, ``The protective action decision model: theoretical modifications and
  additional evidence,'' \emph{Risk Analysis: An International Journal},
  vol.~32, no.~4, pp. 616--632, 2012.

\bibitem{terpstra_citizens_2013}
T.~Terpstra and M.~K. Lindell, ``Citizens’ perceptions of flood hazard
  adjustments: an application of the protective action decision model,''
  \emph{Environment and Behavior}, vol.~45, no.~8, pp. 993--1018, 2013.

\bibitem{scovell_north_2018-1}
M.~Scovell, C.~McShane, A.~Swinbourne, and D.~Smith, ``North {Queenslanders}'
  perceptions of cyclone risk and structural mitigation intentions. {Part} {I}:
  psychological and demographic factors,'' Jul. 2018.

\end{thebibliography}

\end{document}